\documentclass[final]{article}
\usepackage[square,numbers]{natbib}

\usepackage{neurips_2023}

\usepackage[utf8]{inputenc} 
\usepackage[T1]{fontenc}    
\usepackage{hyperref}       
\usepackage{url}            
\usepackage{booktabs}       
\usepackage{amsfonts}       
\usepackage{nicefrac}       
\usepackage{microtype}      
\usepackage{xcolor}         
\usepackage{float}

\usepackage{multirow}

\usepackage{sidecap}
\usepackage{amsthm}
\usepackage{amsmath}
\usepackage{amssymb}
\usepackage{amsfonts}
\usepackage{verbatim} 
\usepackage{url}
\usepackage{pifont}
\usepackage{eqparbox}
\usepackage{arydshln} 
\usepackage{bigstrut}
\usepackage{comment}
\usepackage[linewidth=1pt]{mdframed}
\usepackage{framed}
\usepackage{xcolor,colortbl}
\usepackage{pgffor}
\usepackage{comment}
\usepackage{graphicx}
\usepackage{booktabs}
\usepackage{subcaption}
\usepackage{enumitem}
\usepackage{makecell}
\usepackage{listings}
\usepackage{cleveref}
\usepackage{soul}

\usepackage{boxedminipage}
\usepackage{xspace}
\usepackage{array}
\usepackage{framed}
\usepackage{caption}
\usepackage{tikz}
\usetikzlibrary{tikzmark}
\usepackage{enumitem}
\usepackage{bigstrut}
\usepackage{placeins}
\usepackage[notextcomp]{stix}
\usepackage{wrapfig}

\newcommand{\tulu}{\textsc{T\"ulu}}

\newcommand{\llama}{\textsc{LLaMa}}

\definecolor{lightergray}{RGB}{230,230,230}
\definecolor{DarkRed}{RGB}{130,25,0}
\definecolor{DarkGreen}{RGB}{30,130,30}
\definecolor{DarkBlue}{RGB}{0,0,250}
\definecolor{purple}{rgb}{0.5,0,1}
\definecolor{dcyan}{rgb}{0.2,0.6,0.5}
\definecolor{darkgreen}{rgb}{0,200,0}
\definecolor{light-gray}{gray}{0.95} 
\definecolor{darkred}{RGB}{200,0,0}
\definecolor{lightgreen}{RGB}{231,255,219}
\definecolor{lightred}{RGB}{252,231,234}
\definecolor{lightyellow}{RGB}{250,253,191}

\title{Parameter Efficient Instruction Tuning: \\ An Empirical Study} 
\author{%
Pengfei He\\
AdaBit AI
}

\begin{document}

\maketitle

\begin{abstract}
Instruction tuning \citep{zhang2023instruction} \citep{wei2022finetuned} \citep{ouyang2022training} \citep{wang2023selfinstructaligninglanguagemodels} has become an important step for finetuning pretrained language models to better follow human instructions and generalize on various tasks. Nowadays, pretrained language models\citep{bommasani2022opportunities} become increasingly larger, and full parameter finetuning is overwhelmingly costly. Therefore, Parameter Efficient Finetuning (PEFT) \citep{chen2023parameterefficient} \citep{liu2022fewshot} \citep{he2022unified} has arisen as a cost-effective practice for instruction tuning because of significantly smaller computational, memory, and storage cost compared to full finetuning. Despite their widespread adaptations, the vast hyperparameter spaces, the number of PEFT methods, the different focus of instruction tuning capabilities make disentangling the impact of each aspect difficult. This study systematically investigates several representative PEFT methods, surveying the effect of \textbf{hyperparameter choices} including training hyperparameters and PEFT-specific hyperparameters, how different \textbf{models sizes} and \textbf{the number of instruction tasks} affect the performance, \textbf{in-task-distribution memorization} and \textbf{open instruction following capability}\citep{wang2023far}. 
Our empirical study shows that only LoRA~\cite{hu2021lora} and adapter~\citep{houlsby2019parameterefficient} can get close to full finetuning with ideal training settings. The ideal training setting includes an appropriate learning rate, largest LoRA rank or adapter size allowed and a diverse training tasks. On the other hand, LoRA and adapter suffer from training instability if such an ideal training condition is not met. Additionally, LoRA requires a greater number of tasks for effective unseen task generalization, exhibit slower learning speed. Moreover, LoRA has weaker task-level memorization. Lastly, LoRA and adapter falls short in complex reasoning, coding and long-form generation compared to finetuning in open instruction tuning settings but it shows stronger capabilities compared to adapter.
We hope our work could guide practitioners through the PEFT optimization and provides the insight of further research on what aspects to improve these methods in instruction tuning scenarios.

\end{abstract}

\section{Introduction}
\label{sec:intro}
 Recent study demonstrates instruction tuning which trains language models on instruction-output pairs data can enhance model's capability of comprehending human instructions and following these instructions. Efforts are also underway to compile a diverse mixture of high-quality instruction tuning data\citep{wang2022super} \citep{wang2023far}\citep{ivison2023camels} \citep{zhou2023lima} to advance language models in their better generalization on downstream tasks and better alignment with user intents. Consequently, instruction tuning has become a standard method of aligning LLMs closely with human instructions. However, LLM's vast amounts of parameters often constrains their accessibility through traditional finetuning methods due to the high training, storage, memory cost it incurs. In light of this, PEFT has recently demonstrated remarkable achievements to address the concern. This is primarily due to its cost-efficiency, as PEFT necessitates the training of only a fraction of the model's parameter, and it results in much lower required training memory and storage.

Adapter\citep{houlsby2019parameterefficient}, a pinoeer PEFT method, is a bottleneck network inserted between layers within fixed pretrained model. LoRA \citep{hu2021lora} trains on a low-rank weight matrix, applied additively to selected matrix independently within the transformer layer.
\citet{zaken2022BitFit} proposes to finetune the biases of the neural network only. 
Inspired by the effectiveness of text prompting methods in directing LLM, Prompt Tuning \citep{lester2021power} and Prefix Tuning \citep{li2021prefixtuning} have been developed, they concatenate a sequence of soft tokens(continuous vectors) into the model input or activation and only train these soft tokens.

In light of these, our study begins by identifying effective PEFT methods. This involves training on expert-written instruction tuning dataset SuperNI\citep{wang2022super} and conducting a series of experiments. These experiments vary in hyperparameters, data sizes, data distribution and model sizes. To explore the PEFT's effectiveness further, we extend our training and evaluation to a more challenging setup \tulu \citep{wang2023far}, and it consists of more diverse instruction tuning tasks and more comprehensive evaluation which covers a range of complex model capabilities (i.e., factual knowledge, reasoning, multilinguality, coding) and open-ended instruction-following abilities.

Our key findings are as follows:
\begin{enumerate}
    \item We identify LoRA and adapter are most effective PEFT methods for instruction tuning (Sec.  \ref{subsec:method_comparison}). Greater LoRA ranks/adapter sizes help improving performance (Sec. \ref{subsec:larger_rank}).
    \item We find that LoRA and adapter training exhibits certain instability compared to finetuning, and such an instability correlates with greater rank and higher learning rate (Sec. \ref{subsec:lora_lr_sensitive}).
    \item We validate that PEFTs with larger base models consistently improve the final performance (Sec. \ref{subsec:lora_base_model}).
    \item We show that both LoRA and adapter have weaker generalization capability in low-data instruction tuning settings (Sec. \ref{para:lora_low_data}), and LoRA shows weaker task-level memorization when compared to both adapter and finetuning (Sec. \ref{subsec:task_level_mem}).
    \item For open instruction tuning, LoRA has generally better multifaceted capabilities than adapter and make it an ideal alternative to finetuning. Plus, both adapter and LoRA demonstrates weak coding, complex reasoning and long-form generation capabilities (Sec. \ref{subsec:open_instruct}).
\end{enumerate}

\section{Experiment Setups}

\subsection{Setup 1: T5 finetuning on SuperNI}
\label{subsec:superni_setup}
In the first experiment, our goal is to identify the instruction following capabilities of PEFTs. We limit the hyperparameter search space along learning rates and PEFT-specific hyperparameter only. We choose to use SuperNI \citep{wang2022super} as our dataset for two reasons. First, it comprises of a large amount of different tasks, offering an ideal testbed for the cross-task generalization capabilities of LLMs. Second, compared to some latest larger instruction data mixture, it's considerably smaller which significantly reduces per-experiment runtime and thus overall runtime greatly. For the first setup, we use model T5-3B "LM-adapted" language model \citep{raffel2020exploring} which is further trained with a language modeling objective. 

In our standard setup, our full dataset comprises 707 training tasks and 50 validation tasks, all randomly selected from SuperNI \citep{wang2022super} original training tasks. The test data compromises 119 unseen English tasks adhering the original setup. For each task, we sample 100 instances. In preparing the data examples aligning with instruction data format, we concatenate the task definition and the task input for each data example. Based on the standard setup, we vary model sizes, data sizes, PEFT for ablation study of the correlation of one or more variables.

Each PEFT method is trained for 4 epochs to ensure adequate training for each PEFT method. For PEFTs, we utilize the AdamW optimizer, paired with a linear learning rate scheduler and a warm-up phase that comprises 3\% of the total training steps. For finetuning, we use constant learning rate scheduler which kept the same training setting as Tk-Instruct \citep{wang2022super}. We use RougeL \citep{lin-2004-rouge} score as our evaluation metric, and the best-performing checkpoint is selected for evaluation on the test set. Overall, our first experiment design represents a balanced trade-off between testing effectiveness and identifying patterns of PEFT and efficiency of running experiments at scale.

Given the heterogeneous nature of PEFT methods, the ratios of their trainable parameters span a wide range, and their PEFT-specific hyperparameter varies by their design space. The search range for PEFT-specific hyperparameters and learning rates can be referred to Table \ref{tab:method_comparison}.

\subsection{Setup 2: \llama-2\ finetuning on \tulu\ datasets}
\label{subsec:tulu_setup}
With the selected PEFTs and their best hyperparameters found from the first setup, we want to compare PEFTs based on modern LLMs with finetuning in more comprehensive instruction training and evaluation settings. For this purpose, we utilize \llama-2 \citep{touvron2023llama} language model as our base model. We leverage open instruction datasets \tulu \citep{wang2023far} which compromises both human-generated and GPT-generated instruction tuning data, and we follow the same multi-faceted evaluation setup covering factual knowledge, reasoning, multilinguality, coding and open-ended instruction following.

We train selected PEFT methods for 3 epochs. These were conducted with a maximum input length 2048, consistent with finetuning setting from \tulu \citep{wang2023far}. Refer to Sec. \ref{app:hyperparams} for detailed training hyperparameters.

\section{Empirical Findings}
\label{sec:findings}
We derived Finding 1 through 6 from Setup \ref{subsec:superni_setup}, and Finding 7 from Setup \ref{subsec:tulu_setup}. Finding 3 through 6 provides a deeper exploration into LoRA.


\subsection{Finding 1: LoRA and adapter are effective for instruction tuning.}
\label{subsec:method_comparison}

\begin{table}[h!]
    \centering
    \caption{Experiments containing the grid search on PEFT hyperparameters and the best results across all hyperparameter combinations. $r$ is the LoRA rank, $\eta$ is the learning rate, $l$ is the prefix/prompt length, $h$ is the reparameterized MLP dimension. For each PEFT method, best performance is reported as the best RougeL score on SuperNI across all hyperparameter combinations.  Additionally, we list the ratio between the trainable PEFT parameters and the model frozen parameters corresponding to the best performance setting.}
    \label{tab:method_comparison}
    \begin{tabular}{llccc}
        \toprule
        Method & Searched hparams & Best hparams & \% trainable & Best Perf. \\
        \midrule[0.8pt]
        LoRA & \makecell[l]{$r \in \{8,32,64,128,256,512\}$,\\ $\eta \in \{1\text{e-}5, 5\text{e-}5, 1\text{e-}4, 5\text{e-}4, 1\text{e-}3 \}$} & \makecell[l]{$r$=512,\\ $\eta$=1e-4} &  9.6\% & 47.1 \\ \hline
        Adapter & \makecell[l]{$s\in \{8,32,64,128,256, 512\}$,\\$\eta \in \{1\text{e-}5, 5\text{e-}5, 1\text{e-}4, 5\text{e-}4, 1\text{e-}3 \}$} & \makecell[l]{$s$=512,\\ $\eta$=1e-4}  & 6.6\% &  46.7  \\ \hline
        Prefix Tuning & \makecell[l]{$l \in \{8, 32, 64, 128,256, 512\}$,\\$h=\{null, 256, 512,1024\}$,\\$\eta \in \{1\text{e-}5, 5\text{e-}5, 1\text{e-}4, 5\text{e-}4, 1\text{e-}3 \}$} & \makecell[l]{$l$=512,\\ $h$=256,\\ $\eta$=1e-4}  & 0.9\% &   41.6 \\ \hline 
        
        Prompt Tuning & \makecell[l]{$l \in \{8,32,128,256 \}$,\\$\eta \in \{1\text{e-}5, 5\text{e-}5, 1\text{e-}4, 5\text{e-}4, 1\text{e-}3 \} $} & \makecell[l]{$p$=8,\\ $\eta$=5e-5}  & 0.001\% &  16.3 \\ \hline 
        
        BitFit & \makecell[l]{$\eta \in \{1\text{e-}5, 5\text{e-}5, 1\text{e-}4, 5\text{e-}4, 1\text{e-}3 \}$} & \makecell[l]{$\eta$=5e-4} & 0.04\% & 41.5\\
        
        \midrule[0.8pt]
        Full Finetuning & $\eta \in \{1\text{e-}5, 5\text{e-}5, 1\text{e-}4, 5\text{e-}4, 1\text{e-}3 \}$ & \makecell[l]{$\eta$=1e-5}  & 100\% & 47.8 \\ 
        \bottomrule
    \end{tabular}
\end{table}

Our extensive hyperparameter search reveals that among the five PEFTs, only LoRA and adapter are proved close to full finetuning in instruction tuning settings (See Table \ref{tab:method_comparison}). Prompt tuning shows no effective learning due to the difficulty of optimizing soft prompts as claimed in some other findings \citep{hu2021lora} and the cross-task nature of instruction tuning. Prefix tuning shows some learning but still significantly underperforms finetuning. This observation is in line with recent theoretical analyses\citep{petrov2023prompting}, which suggests that prefix tuning and prompt tuning are less expressive than finetuning. Prefix tuning also has a reparameterization trick that transforms the prefix matrix into MLP, and our experiment results indicate such reparameterization improves training stability. Lastly, BitFit also falls short compared to finetuning, and we think tuning bias alone also has limited expressivity.

The detailed experimental results for each PEFT are attached in the Sec. \ref{sec:peft_hp_search}.

\subsection{Finding 2: More PEFT trainable parameters yield better performance.}
\label{subsec:larger_rank}

\begin{figure}[H]
    \centering
    
    \begin{subfigure}{0.49\textwidth}
        \centering
        \includegraphics[width=\linewidth]{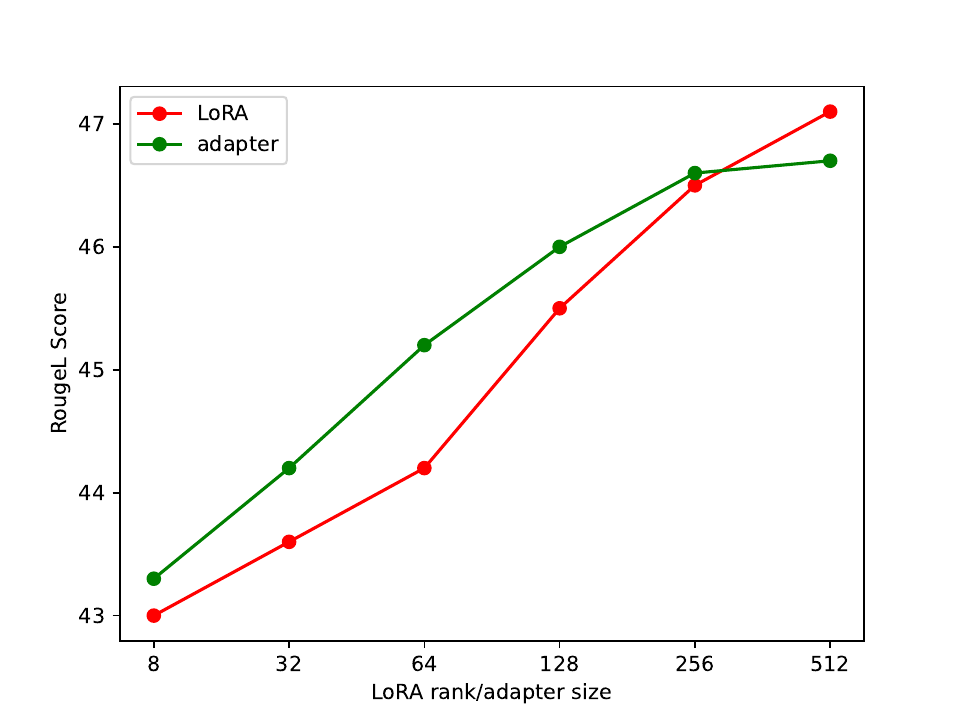}
        \caption{Performance of different LoRA ranks and adapter size. Each point represents averaged score across three runs with different random seeds.}
        \label{fig:trainable_parameter}
    \end{subfigure}
    \hfill
    \begin{subfigure}{0.49\textwidth}
        \centering
        \includegraphics[width=\linewidth]{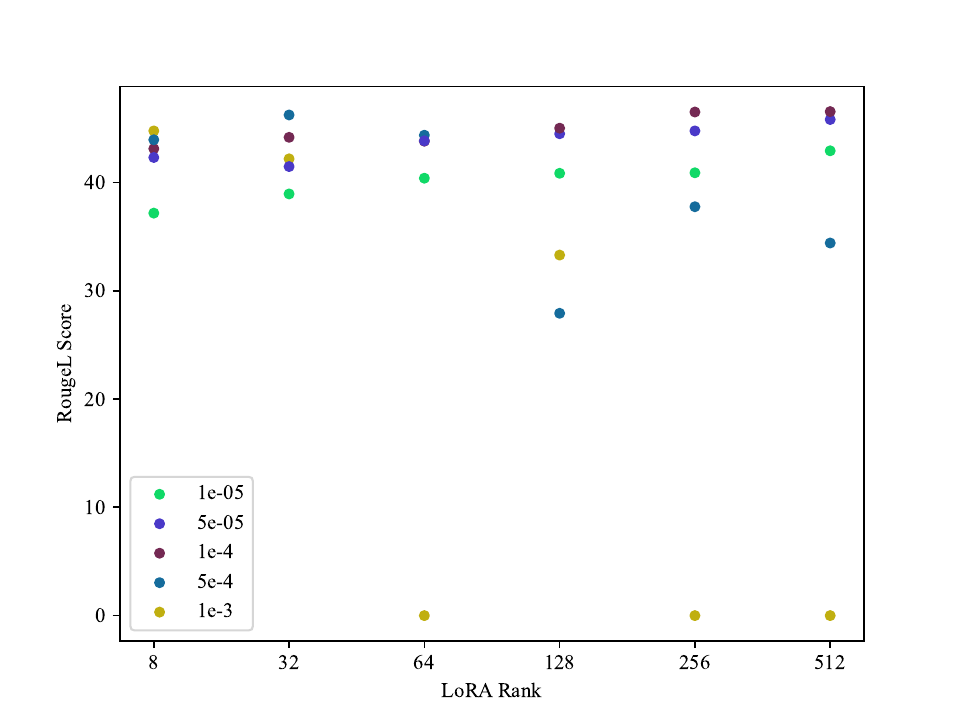}
        \caption{Performance of different LoRA ranks and learning rates. Legends show different learning rates. Each point represents a run with different random seeds.}
        \label{fig:lora_lr}
    \end{subfigure}
    \caption{The impact of LoRA ranks and adapter size on LLM performance}
    \label{fig:data_size}
\end{figure}

We have investigated the impact of LoRA rank and adapter size in instruction tuning. As indicated by Fig. \ref{fig:trainable_parameter}, we observe that a higher LoRA rank/adapter size consistently yields better performance when coupled with the optimal learning rate. The finding aligns with the general principle in machine learning, where an increase in the number of parameters correlates with enhanced model capacity and performance improvements. Also, due to the nature of task diversity in instruction tuning settings, more tasks coupled with more PEFT trainable parameters do improve model performance. However, the performance gain coupled with higher rank tends to diminish, despite the rank increases by the factor of two. See detailed results in Table \ref{tab:peft_hp_search_lora} and Table \ref{tab:peft_hp_search_adapter}.

\subsection{Finding 3: Higher LoRA rank is more sensitive to the learning rate}
\label{subsec:lora_lr_sensitive}

As Fig. \ref{fig:lora_lr} indicates, too small learning rate could cause underfitting given the same training epochs while high learning rate could cause training instability. 
If we also take rank into consideration, a lower rank has a better tolerance of high learning rate. For example, at rank 8, learning rate 1e-3 even produces the best result at the same rank. However, as the rank goes higher, relatively lower learning rate stabilize training and improves performance. According to the grid search result, the optimal learning rate is 1e-4 among experiments with different LoRA ranks.



\subsection{Finding 4: LoRA and adapter with larger base models perform better}
\label{subsec:lora_base_model}
\begin{figure}[H]
    \centering
    \begin{subfigure}{0.49\textwidth}
        \centering
        \includegraphics[width=\linewidth]{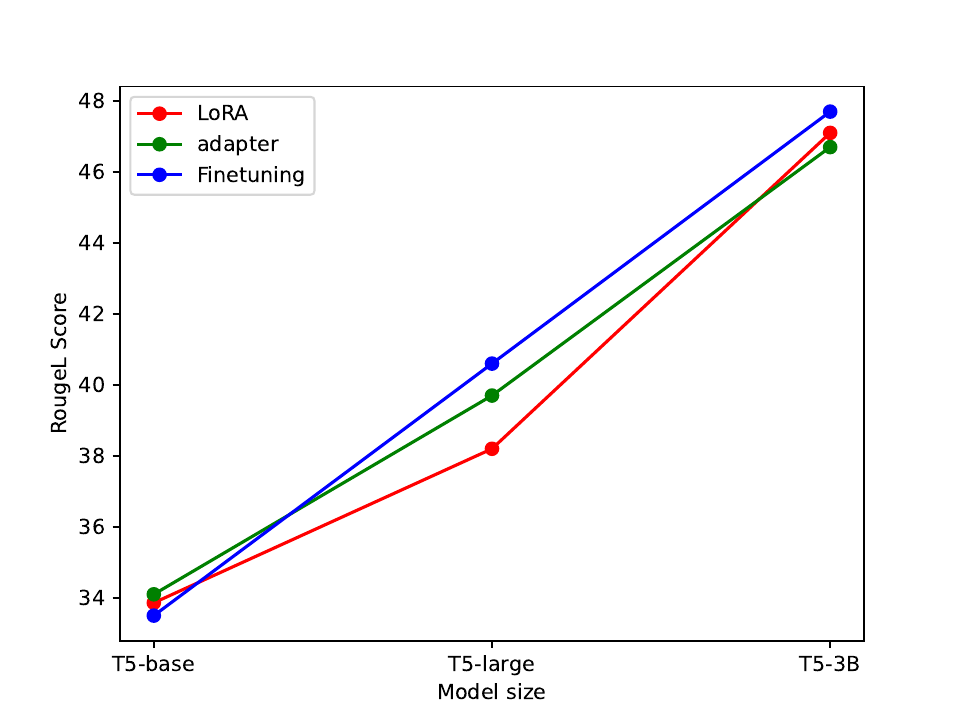}
        \caption{Performance of different model sizes. Legends show tuning methods. Each point represents a run with different random seeds.}
        \label{fig:model_size_peft}
    \end{subfigure}
    \hfill
    \begin{subfigure}{0.49\textwidth}
        \centering
        \includegraphics[width=\linewidth]{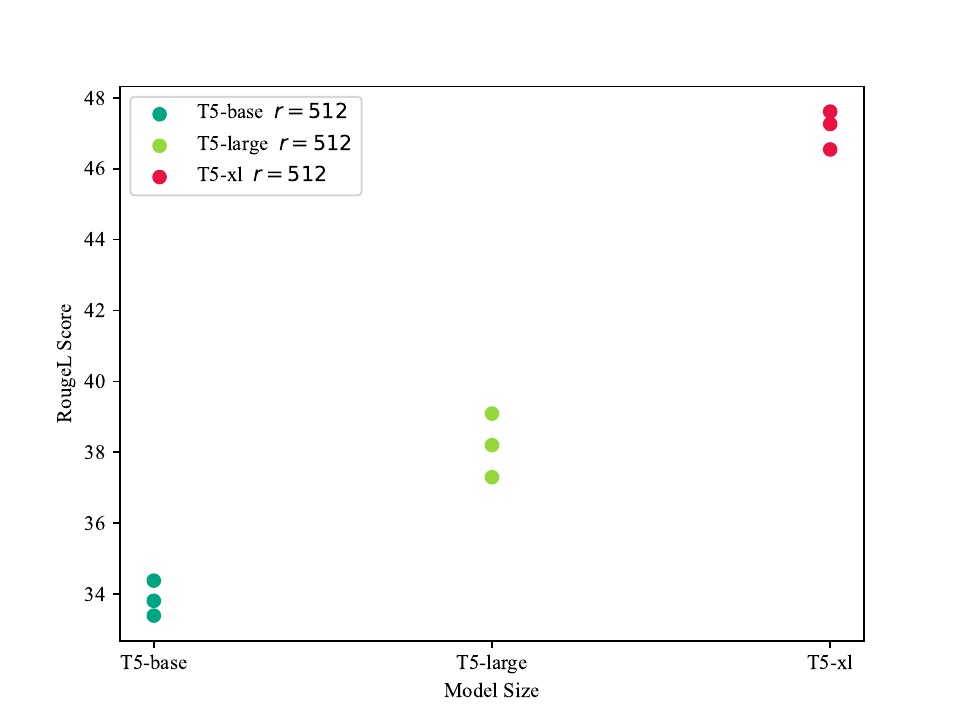}
        \caption{Performance of LoRA under different model size. Each point represents a run with different random seeds.}
        \label{fig:lora_model_size}
    \end{subfigure}
    \caption{Model size impact on performance.}
    \label{fig:model_size}
\end{figure}

Both Fig. \ref{fig:model_size_peft} and Fig. \ref{fig:lora_model_size} illustrates that the performance of both LoRA and adapter is consistently improved from T5-base to T5-3b.  This trend observed indicates that as the capacity of the underlying model increases, the ability to finetune LoRA and adapter also becomes more effective. Consequently, this highlights the importance of selecting an appropriately scaled model to maximize the benefits of LoRA.



\subsection{Finding 5: LoRA underperforms in low-data setting}
\label{para:lora_low_data}
\begin{figure}[H]
    \centering
    \begin{subfigure}{0.49\textwidth}
        \centering
        \includegraphics[width=\linewidth]{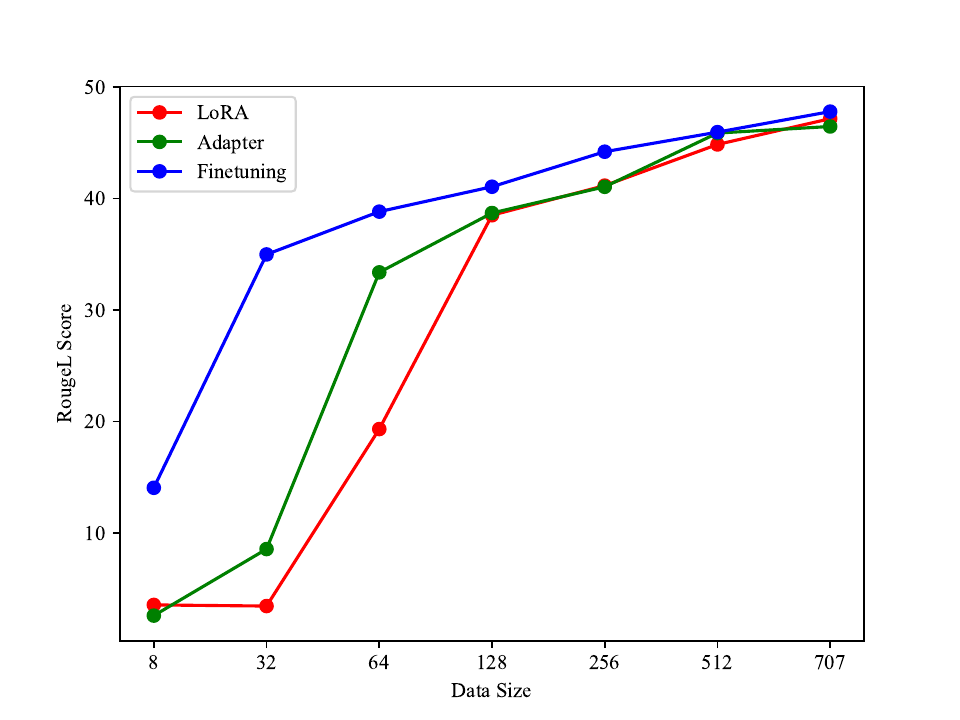}
        \caption{Performance of LoRA, adapter and finetuning under different data sizes. All hyperparameters are set to be optimal according to Table \ref{tab:method_comparison}. RougeL scores are averaged across three runs with different random seeds.}
        \label{fig:data_size_n_methods}
    \end{subfigure}
    \hfill
    \begin{subfigure}{0.49\textwidth}
        \centering
        \includegraphics[width=\linewidth]{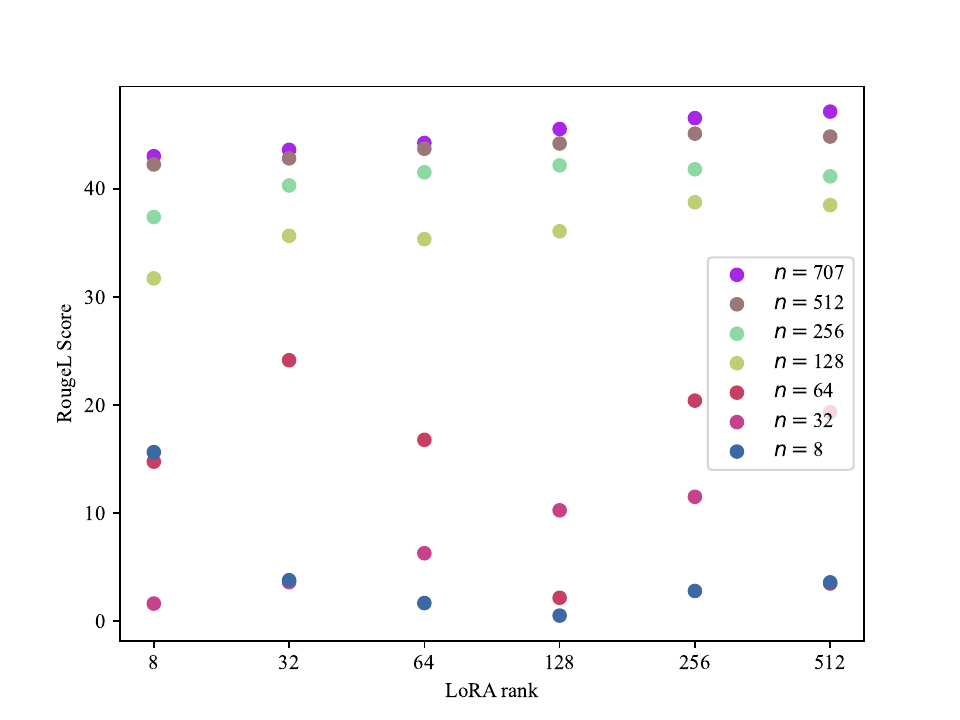}
        \caption{Performance of LoRA under different data sizes and LoRA ranks. $n$ is the number of training tasks. Each point represents a run with different random seeds.}
        \label{fig:lora_data_size}
    \end{subfigure}
    \caption{Data size impact on LLM performance.}
    \label{fig:data_size}
\end{figure}

Despite fully trained LoRA is close to or on par with full finetuning, larger data size always benefits LoRA across ranks as shown in Fig. \ref{fig:lora_data_size}. On the other hand, in many industrial adaptations, there could only be a limited number of instruction tasks available. The important question is how many different instruction tasks required so that it starts to generalize for each method? In our experiments, we reveal that LoRA is actually a "slow learner" in a way that it requires more tasks to ramp up cross-task generalization than adapter and finetuning (See Fig. \ref{fig:data_size_n_methods}), and it suffers from training instability when the number of training tasks is low (See Fig. \ref{fig:lora_data_size}). Therefore, in low-data instruction tuning settings, especially domain specific, finetuning is still an optimal choice, and adapter could be a PEFT alternative if the slight overhead latency is acceptable.



\subsection{Finding 6: LoRA has worse task-level memorization}
\label{subsec:task_level_mem}
Under the ideal downstream tuning setting, it is advantageous to provide training instruction tasks same as testing instruction tasks, and this resemble traditional training and test dataset split on a instance level. This raises a natural research question: "\textit{How do LoRA and adapter perform for test data which is in-distribution in task level but out of distribution in instance level?}" In such scenarios, a key capability of language models is their memorization ability: how effectively they can learn and retain task-level features seen during training and perform well on tasks of the same types during testing. To assess this, we selected 100 tasks instances from each training task type, provided there were sufficient instances outside of training set. Our experiments indicates that LoRA demonstrates comparatively weaker task-level memorization capabilities than both adapter and finetuning. By contrast, adapter is just slightly weaker than finetuning in this aspect. (Fig. \ref{peft_traditional_test})
We hypothesize that the key reason for memorization capability lies in the number of extra parameters with nonlinearities, as pre-trained knowledge is predominantly located in the feed-forward network(FFN) layers \citep{dai2022knowledge} rather than attention layers. Consequently, since there is no additional nonlinear layer for LoRA and only query and key projection layer weight delta are tuned in our LoRA configuration, this results in a reduced capability to store task-specific knowledge.

It's worth noting that the study by Mireshghallah et al. \citep{mireshghallah2022memorization} investigate instance-level memorization to reduce extraction attack, and suggests that the adapter model exhibits comparatively less memorization at this level. In addition to our finding, we consider adapter could be a valuable alternative if both task-level in-distribution memorization and privacy are important for downstream tasks despite the slight inference latency.

\begin{table}[h!]
    \centering
    \caption{Performance of LoRA, adapter and fine-tuning tested on training tasks but unseen instances on SuperNI.}
    \label{tab:peft_traditional_test}
        \begin{tabular}{lccccc}
        \toprule
        & Rouge-L & \\
        \midrule
        LoRA ($r=512$) & 55.4  & \\
        Adapter ($s=512$) & 58.4 &  \\
        Fine tuning & 59.7 &  \\
        \bottomrule
    \end{tabular}
\end{table}



\subsection{Finding 7: PEFTs underperform in open instruction tuning and multifaceted testing}
\label{subsec:open_instruct}


As Table \ref{tab:peft_tulu} suggests, both LoRA and adapter exhibits a significantly weaker performance in reasoning tasks while LoRA consistently outperforms adapter when it comes to a wide range of challenging open-instruction tasks.

\begin{table}[h!]
    \centering
    \small
    \setlength{\tabcolsep}{4pt} 
    \caption{Performance of LoRA($r=512$), adapter($s=512$) and finetuning based on \llama-2 7B and trained on \tulu-1.1 data mixture. $r$ is LoRA rank and $s$ is adapter size.}
    \label{tab:peft_tulu}
        \scalebox{0.9}{
            \begin{tabular}{lcccccc}
            \toprule
            & MMLU & GSM & BBH & TydiQA & Codex-Eval & \multirow{2}{*}{Average}  \\
            & (factuality) & (reasoning) & (reasoning) & (multilinguality) & (coding) & \\
            \cmidrule(){2-7}
             & EM & EM & EM & F1 & P@10 &  \\
             & (0-shot) & (8-shot, CoT) & (3-shot, CoT) & (1-shot, GP) & (0-shot)   \\
             
            \midrule
            LoRA  & 49.7 & 29.1 & 43.3 & 52.1  & 19.7 & 38.78\\
            Adapter  & 46.9 & 20.5 & 40.8  & 48.4 & 19.7 & 35.26 \\
            Finetuning & 49.2  & 37.0 & 44.2 &  52.9 & 33.9  & 43.4 \\
            \bottomrule
        \end{tabular}
        }
\end{table}

\clearpage
\section{Related Work}
\label{sec:related}

Large Language Models (LLMs) have revolutionized the field of natural language processing with their vast knowledge base and advanced reasoning capabilities. Yet, their extensive parameter sizes pose challenges for traditional downstream finetuning.

As summarized in \citet{lialin2023scaling}, there are five distinct categories of PEFT methods with some methods straddling multiple categories. For the initial stage of our survey, we have selected seven PEFT method that span these categories. We employ LoRA \citep{lester2021power} as a representative of reparameterization-based method, building on its proven effectiveness in works such as QLoRA \citep{dettmers2023qLoRA}. BitFit \citep{zaken2022BitFit} serves as our selective method. Prompt-Tuning \citep{lester2021power} and Prefix-Tuning \citep{li2021prefixtuning} are selected from the intersection of soft prompts and additive methods. Lastly, we include adapter \citep{houlsby2019parameterefficient} method which overlaps both additive and adapter-based method.

The line of research related to instruction tuning has developed in several key ways.
\citet{ziegler2020finetuning} focused on finetuning language models through the use of human preferences, aiming to produce model outputs that better align with human intent and values. SuperNI \citep{wang2022super} introduced an expert-written instruction dataset that spans a variety of task types. \citet{wang2022selfinstruct} leveraged pre-trained models to autonomously generate instructional data for subsequent finetuning. This approach led to notable enhancements in the model's capability to accurately follow instructions, mitigating the need for costly human-annotated instruction datasets. \citet{alpaca} contributed a synthetic dataset generated from GPT-4 outputs. This dataset was created using self-instruct methods and was subsequently distilled for the purpose of instruction tuning. \citet{wang2023far} studies how the combination of human annotated and GPT4 generated instruction dataset modulate the performance of trained model, with a blend of both proving optimal. In our experiment, we utilize the same experiment setting but with the integration of PEFT methods.

As reported in \citet{biderman2024loralearnsforgets} LoRA learns less but has better regularization effect, and it is sensitive to learning rates. Our finding is complementary to theirs with more focus on instruction tuning. In subsection \ref{subsec:task_level_mem}, LoRA also has a weaker in-task-distribution memorization capability compared to finetuning, and in subsection \ref{para:lora_low_data} LoRA requires more instruction tasks to generalize. 
Different aspects of PEFTs for instruction tuning have also been explored. \citet{zhuo2024astraiosparameterefficientinstructiontuning} investigates PEFTs for instruction tuning on coding, and it reveals that full finetuning still surpass all PEFTs in terms of downstream performance.
Plus, \citet{biderman2024loralearnsforgets} also tests coding and math reasoning capability with domain dataset, and it shows LoRA is strong at math but weaker at coding. In subsection \label{subsec:open_instruct}, with mixed-task dataset in our work, we find LoRA shows inferior performance in a wider range of challenging tasks including coding, complex reasoning and open-ended generation by compared to finetuning.






\section{Conclusion}
This study has demonstrated that PEFTs, particularly LoRA and adapter, present viable alternatives to full fine-tuning in instruction-tuning scenarios, offering a balance between performance and computational efficiency. Our comprehensive empirical analysis highlighted that greater LoRA ranks and adapter sizes enhance performance significantly, though they may introduce some training instability. Additionally, while LoRA outperforms adapter in open instruction tuning settings due to its robust generalization across diverse tasks, it requires a substantial number of tasks to achieve effective unseen task generalization.

Moreover, the limitations observed in the long-form generation capabilities of both LoRA and adapter underscore the ongoing challenges within PEFT methods, necessitating further innovation and exploration in this field. These findings advocate for a nuanced application of PEFT methods, tailored to specific model sizes, task types, and data availability, aligning them more closely with real-world applications. As the landscape of language model finetuning evolves, the insights from this study will hopefully guide future research towards optimizing the efficiency and effectiveness of instruction tuning across various domains.

Future work should focus on refining these methods to enhance their stability and expand their applicability to more complex and varied datasets. By continuing to investigate the trade-offs between different PEFT strategies and their impacts on model performance, the field can move towards more sophisticated and nuanced PEFT  techniques that maximize both performance and efficiency.

\bibliographystyle{abbrvnat}
\bibliography{main
}

\clearpage
\appendix
\begin{center}
{\Large \textbf{Supplementary Material}}
\end{center}

\begin{ack}
We thank our founding advisor Jieyu Zhang at AdaBit AI for proofreading this paper.
\end{ack}

\section{Limitations}
Despite the comprehensiveness of our training and evaluations, we do not exhaustively cover all PEFT methods and more fine-grained hyperparameter grid search. In our work, we only select the most representative PEFT methods across the broad PEFT categories, and commonly used hyperparameters.

Given the computing constraint, our first set of experiments about hyperparameter search is based on SuperNI and T5. Therefore, the optimal hyperparamter based on them might not reflect latest model architecture's performance on latest instruction tuning datasets which has a broader coverage on different topics. Plus, \tulu \ suggests incorporating SuperNI in the data mixture is harmful for model performance.

\section{Broader Impact}
We believe that a comprehensive validation of a PEFT method is broadly positive. And it could help practitioners to narrow the search space of PEFT methods and hyperparameters to save experiment time and efforts.


\section{Model Training Details and Compute}
\label{app:hyperparams}

We designed the first set of experiments to find the hyperparameter pattern with the considerations of each PEFT's feature.
For instance, the prompt length in prompt tuning \citep{lester2021power} and the prefix length in prefix tuning \citep{li2021prefixtuning} are both constrained by the input length. BitFit \citep{zaken2022BitFit} features significantly fewer trainable parameters ($\ll 1\%$) and lacks any adjustable PEFT hyperparameters. Considering these complexities, we opted not to standardize the number of trainable parameters for performance comparison. Instead, our experimental design adheres closely to the original configurations as reported in the respective foundational works. We select these representative PEFTs and conducted experiments within a restricted hyperparameter search space to yield meaningful insights.

For SuperNI \citep{wang2022super} experiments in Sec. \ref{subsec:superni_setup}, we trained our models primarily on High-Flyer cluster, each node on which contains 8 Nvidia A100 GPUs. We utilize DistributedDataParallel for most training jobs when GPU memory permits; otherwise, we employ the DeepSpeed library and ZeRO optimizer. Our training hyperparameters that are not part of the grid search are as follows:

\begin{itemize}
    \item Precision: FP32
    \item Epochs: 4
    \item Weight decay: 0
    \item Warmup ratio: 0.03
    \item Max. seq. length: 1,024
    \item Effective batch size: 128
    \item Dropout: 0.1
    \item LoRA Layers wrapped: all query and key layers
\end{itemize}

For \tulu \ experiments in Sec. \ref{subsec:tulu_setup}, we trained our models primarily on a local cluster, each experiment is conducted on a single NVIDIA A6000 GPU without using extra training framework. Our training hyperparameters are as follows:
\begin{itemize}
    \item Precision: BFloat16
    \item Epochs: 3
    \item Weight decay: 0
    \item Warmup ratio: 0.03
    \item Learning rate: 1e-4
    \item Max. seq. length: 4,096
    \item Effective batch size: 128
    \item LoRA Rank: 512
    \item LoRA Alpha: 512
    \item LoRA dropout: 0.1
    \item Layers wrapped: all query and key layers
\end{itemize}

\section{Reproducibility}

We cannot guarantee the exact reproduction of all experiment numbers because the experiments are conducted on preemptible clusters, which may restart multiple times. However, the overall findings remain valid. Plus, there are some experiment runs with unstable training failed to generate non-empty content, and it leads to near-zero RougeL scores and reduced average RougeL scores reported in the results. Such results are caused by the nature of training instability of PEFTs, high learning rates or a large volume of trainable parameters.




\section{PEFT hyperprameter search results}
\label{sec:peft_hp_search}
\begin{table}[h!]
    \centering
    \caption{Performance of LoRA with different hyperparameters. $r$ denotes the LoRA rank.}
    \label{tab:peft_hp_search_lora}
    \begin{tabular}{lcccccc}
        \toprule
         & \multicolumn{5}{c}{lr} \\
        \cmidrule(lr){2-6}
        & 1e-05 & 5e-05 & 1e-4 & 5e-4 & 1e-3 \\
        \midrule
    $r=8$ & 37.2 & 42.3 & 43.0 & 44.3 & 44.8 \\
    $r=32$ & 38.9 & 41.5 & 43.6 & 45.5 & 42.2 \\
    $r=64$ & 40.4 & 43.8 & 44.2 & 44.4 & 27.8 \\
    $r=128$ & 40.8 & 44.5 & 45.5 & 35.1 & 33.3 \\
    $r=256$ & 40.9 & 44.8 & 46.5 & 38.0 & 0.0 \\
    $r=512$ & 42.9 & 45.8 & \textbf{47.1} & 23.2 & 0.0 \\
\bottomrule
\end{tabular}
\end{table}
\begin{table}[h!]
    \centering
    \caption{Performance of adapter with different hyperparameters. $s$ is the adapter bottleneck size.}
    \label{tab:peft_hp_search_adapter}
    \begin{tabular}{lccccc}
        \toprule
         & \multicolumn{5}{c}{lr} \\
        \cmidrule(lr){2-6}
         & 1e-05 & 5e-05 & 1e-4 & 5e-4 & 1e-3 \\
        \midrule
        $s=8$   & 37.3 & 43.9 & 43.3 & 43.2 & 44.9 \\
        $s=32$  & 41.5 & 42.0 & 44.2 & 46.0 & 43.2 \\
        $s=64$  & 41.8 & 44.2 & 45.2 & 20.7  & 5.3 \\
        $s=128$ & 41.1 & 45.1 & 46.0 & 5.8  & 41.4  \\
        $s=256$ & 41.4 & 45.1 & 46.4 & 42.7 & 4.3  \\
        $s=512$ & 42.7 & 46.2 & \textbf{46.7} & 5.0 & 4.3  \\

        \bottomrule
    \end{tabular}
\end{table}

\begin{table}[h!]
    \centering
    \caption{Performance of prompt tuning with different hyperparameters. $l$ is the prompt length. }
    \label{tab:peft_hp_search_prompt_tuning}
        \begin{tabular}{lccccc}
        \toprule
         & \multicolumn{5}{c}{lr} \\
        \cmidrule(lr){2-6}
        & 1e-5 & 5e-5 & 1e-4 & 5e-4 & 1e-3 \\
        \midrule
        $l=8$ & 16.3 & \textbf{16.3} & 16.3 & 16.3 & 16.3 \\
        $l=32$ & 16.3 & 16.3 & 16.3 & 16.1 & 16.2 \\
        $l=128$ & 15.6 & 15.6 & 15.6 & 15.5 & 15.2 \\
        $l=256$ & 15.2 & 15.2 & 15.2 & 15.1 & 9.0 \\
        \bottomrule
    \end{tabular}
\end{table}

\begin{table}[h!]
    \centering
    \caption{Performance of prefix tuning with different hyperparameters. $l$ is the prefix length. $h$ is the hidden dimension of reparameterized feedforward neural network. $h=\text{null}$ indicates the reparameterization is not applied.}
    \label{tab:peft_hp_search_prefix_tuning}
    \begin{tabular}{lccccc}
    \toprule
        & \multicolumn{5}{c}{lr} \\
        \cmidrule(lr){2-6}
    & 1e-05 & 5e-05 & 1e-4 & 5e-4 & 1e-3 \\
    \midrule
    $l=32,\ h=\text{null}$ & 15.1 & 6.4 & 8.3 & 30.7 & 26.5 \\
    $l=32,\ h=256$ & 26.1 & 31.6 & 31.0 & 31.4 & 22.9 \\
    $l=32,\ h=512$ & 30.5 & 31.5 & 21.4 & 20.1 & 15.2 \\
    $l=32,\ h=1024$ & 29.0 & 21.2 & 30.3 & 21.0 & 8.8 \\
    $l=64,\ h=\text{null}$ & 12.2 & 1.0 & 6.7 & 34.0 & 35.3 \\
    $l=64,\ h=256$ & 37.4 & 41.2 & 40.2 & 39.0 & 0.0 \\
    $l=64,\ h=512$ & 39.5 & 40.2 & 40.7 & 39.8 & 37.2 \\
    $l=64,\ h=1024$ & 41.0 & 41.2 & 41.2 & 36.7 & 9.8 \\
    $l=128,\ h=\text{null}$ & 5.0 & 1.8 & 29.3 & 37.9 & 39.4 \\
    $l=128,\ h=256$ & 27.7 & 30.8 & 30.6 & 22.1 & 21.1 \\
    $l=128,\ h=512$ & 31.4 & 31.6 & 20.5 & 39.8 & 23.9 \\
$l=128,\ h=1024$ & 28.8 & 31.2 & 30.9 & 9.3 & 11.2 \\
$l=256,\ h=\text{null}$ & 8.2 & 15.3 & 33.0 & 31.1 & 38.0 \\
$l=256,\ h=256$ & 29.8 & 30.0 & 34.7 & 30.4 & 31.9 \\
$l=256,\ h=512$ & 26.3 & 30.7 & 31.2 & 25.1 & 16.5 \\
$l=256,\ h=1024$ & 25.4 & 18.3 & 31.5 & 4.2 & 12.9 \\
$l=512,\ h=\text{null}$ & 8.3 & 18.6 & 33.6 & 37.7 & 40.0 \\
$l=512,\ h=256$ & 38.1 & 30.5 & \textbf{41.6} & 41.2 & 8.3 \\
$l=512,\ h=512$ & 38.2 & 37.7 & 39.5 & 38.8 & 16.9 \\
$l=512,\ h=1024$ & 40.6 & 32.8 & 34.1 & 21.6 & 6.5 \\
\bottomrule
\end{tabular}
\end{table}


\begin{table}[h!]
    \centering
    \caption{Performance of bitfit with different hyperparameters. }
    \label{tab:peft_hp_search_bitfit}
    \begin{tabular}{lccccc}
        \toprule
         & \multicolumn{5}{c}{lr} \\
        \cmidrule(lr){2-6}
        & 1e-05 & 5e-05 & 1e-4 & 5e-4 & 1e-3 \\
        \midrule
        All biases &  1.3 & 28.1 & 35.0 & \textbf{41.5} & 20.8 \\
        \bottomrule
    \end{tabular}
\end{table}


\section{Model size and LoRA rank/adapter size}
\begin{table}[h!]
    \centering
    \caption{Performance of LoRA with different model sizes and LoRA ranks.}
    \label{tab:model_size_n_peft_k_lora.tex}
    \begin{tabular}{lcccccc}
        \toprule
        & \multicolumn{6}{c}{$r$} \\
        \cmidrule(lr){2-7}
        model & 8 & 32 & 64 & 128 & 256 & 512 \\
        \midrule
        t5-base-lm-adapt & 30.5 & 31.2 & 31.8 & 32.5 & 33.0 & \textbf{33.9} \\
        t5-large-lm-adapt & 34.1 & 36.6 & 35.1 & 37.8 & 36.3 & \textbf{38.2} \\
        t5-xl-lm-adapt & 43.0 & 43.6 & 44.2 & 45.5 & 46.5 & \textbf{47.1} \\
        \bottomrule
    \end{tabular}
\end{table}


\end{document}